\documentclass[letterpaper, 10pt, conference]{config/ieeeconf}
\IEEEoverridecommandlockouts 
\usepackage{etoolbox}
\makeatletter 
\let\NAT@parse\undefined
\makeatother
\usepackage[numbers]{natbib} 

\usepackage{multirow}
\usepackage[bookmarks=true, colorlinks=false,
    linkcolor=black,
    filecolor=magenta,      
    urlcolor=cyan]{hyperref}
\usepackage{url}
\usepackage{graphicx} 
\usepackage{amsfonts,amssymb,amsmath} 
\usepackage{config/psfig} 
\usepackage{graphics} 
\usepackage{epsfig} 
\usepackage[all]{xy}  
\usepackage{algorithm}
\usepackage{algpseudocode}

\usepackage[labelfont=small,textfont={small}]{caption}
\usepackage{subcaption}
\usepackage{epstopdf} 
\usepackage{balance}
\usepackage{color}

\usepackage{amsmath}

\DeclareMathOperator*{\argmax}{arg\,max}

\usepackage[table,xcdraw]{xcolor}
\usepackage{pgf}
\usepackage{adjustbox}

\usepackage{todonotes}
\usepackage{tabularx}
\usepackage{times}
\makeatletter
\title{YoloTag: Vision-based Robust UAV Navigation with Fiducial Markers}

\begin{document}
\author{Sourav Raxit, Simant Bahadur Singh, and  Abdullah Al Redwan Newaz 
\thanks{The authors are with the Department of Computer Science, University of New Orleans, New Orleans, LA 70148, USA (email: \texttt{ \{sraxit, sbsingh3, aredwann\}@uno.edu) }. }
}

\maketitle

\begin{abstract}
    By harnessing fiducial markers as visual landmarks in the environment, Unmanned Aerial Vehicles (UAVs) can rapidly build precise maps and navigate spaces safely and efficiently, unlocking their potential for fluent collaboration and coexistence with humans.
    Existing fiducial marker methods rely on handcrafted feature extraction, which sacrifices accuracy. On the other hand, deep learning pipelines for marker detection fail to meet real-time runtime constraints crucial for navigation applications. In this work, we propose YoloTag \textemdash a real-time fiducial marker-based localization system. YoloTag uses a lightweight YOLO v8 object detector to accurately detect fiducial markers in images while meeting the runtime constraints needed for navigation. The detected markers are then used by an efficient perspective-n-point  algorithm to estimate UAV states. However, this localization system introduces noise, causing instability in trajectory tracking. To suppress noise, we design a higher-order Butterworth filter that effectively eliminates noise through frequency domain analysis.
    We evaluate our algorithm through real-robot experiments in an indoor environment, comparing the trajectory tracking performance of our method against other approaches in terms of several distance metrics.
\end{abstract}

\section{Introduction}

The capabilities of visual fiducial markers extend beyond just augmented reality to also enable important applications in Human-Robot Interaction. Robots empowered with fiducial marker tracking can precisely determine their position and orientation relative to the camera viewing the markers. This allows robots to localize themselves in the camera's field of view without relying on GPS or other external positioning systems. Nonetheless, the reliability of detecting and differentiating multiple fiducial markers simultaneously is critical for accurate and continuous robot pose estimation. As the robot moves through the environment, the algorithm must track each marker transition seamlessly, correlating the updated camera perspective to the robot's movement.

Vision-based localization involves estimating the location and orientation of a robot using cameras as primary sensors.
Two main approaches are Relative Visual Localization (RVL) and Absolute Visual Localization (AVL)~\cite{couturier2021review}.
RVL techniques like Visual Odometry (VO)~\cite{scaramuzza2011visual,Zhu2018EVFlowNetSO,article1} and Visual Simultaneous Localization and Mapping (VSLAM)~\cite{7946260,7219438,article,8421746} incrementally estimate a robot's ego-motion and map unexplored environments by tracking visual features across frames.
However, they suffer from drift over long durations. Absolute Visual Localization 
(AVL)~\cite{Chiu2013RobustVN,Gruyer2014MapaidedLW} aims to achieve drift-free global localization by matching imagery to geo-referenced maps, but can have limited precision.

Unmanned aerial vehicles (UAVs) can utilize fiducial markers as visual landmarks for localization and mapping, benefiting significantly in controlled, low-texture, and computationally constrained environments~\cite{Amiri2022}.
However, existing fiducial marker systems predominantly depend on manually designed image processing techniques to identify black-and-white patterns, such as ARToolKit~\cite{kato1999}, AprilTag~\cite{Olson2011AprilTagAR,7759617} and ChromaTag~\cite{DeGol2017ChromaTagAC}. These systems often face difficulties in ensuring robust detection when confronted with challenges like occlusion and variations in lighting conditions.
Recent work has focused on improving marker robustness to distortions through deep neural networks~\cite{Hu2018DeepCD,Yaldiz_2021,9773975}.
Novel applications include underwater markers for navigation~\cite{inproceedings} and end-to-end learned marker generation and detection~\cite{peace2021e2etag}.
Algorithmic improvements~\cite{1467495,GarridoJurado2014AutomaticGA,5995544} have also enhanced detection efficiency and scalability.
However, existing fiducial marker-based localization still face limitations in detection speed, accuracy, and false positive rejection, particularly under challenging conditions like low resolution, occlusion, uneven lighting, and perspective distortion.


In this work, we propose YoloTag which harnesses the power of state-of-the-art deep learning object detection techniques through the integration of the high-performing YOLOv8 model. 
YoloTag provides an efficient machine learning architecture that enables robust multi-marker detection and 3D pose estimation across varied real-world imagery from UAVs operating in demanding conditions, while meeting real-time performance requirements.
By training YoloTag on this extensive dataset of real UAV imagery, it can learn to generalize effectively and handle the various complexities encountered during actual deployment. These include dealing with varying marker poses from multiple perspectives, tracking landmarks as the UAV moves and articulates, handling changing illumination conditions, overcoming partial occlusions, and being resilient to sensor noise. YoloTag's efficient, streamlined, end-to-end deep learning architecture seamlessly fuses the information from multiple detected markers into a coherent joint 3D pose estimate. This provides precise, continuous self-localization capabilities tailored specifically for robust real-world robotic operations.

Our key contributions are: 
(i) Employing YOLO v8 for rapid landmark detection and pose estimation, surpassing current methods in processing speed.
(ii) Leveraging multi-marker detection outcomes under various real-world scenarios to accurately detect and estimate poses.
(iii) Addressing perception noise by employing a Butterworth filter to refine the estimated trajectory.
(iv) Thoroughly assessing the system on UAV platforms in indoor settings, confirming its robustness, accuracy, and suitability for real-world applications.



\section{Realtime Fiducial Marker Detection using Deep Learning}

\begin{figure}[htb]
\vspace{-12pt}
    \centering
    \includegraphics[width=0.5\textwidth]{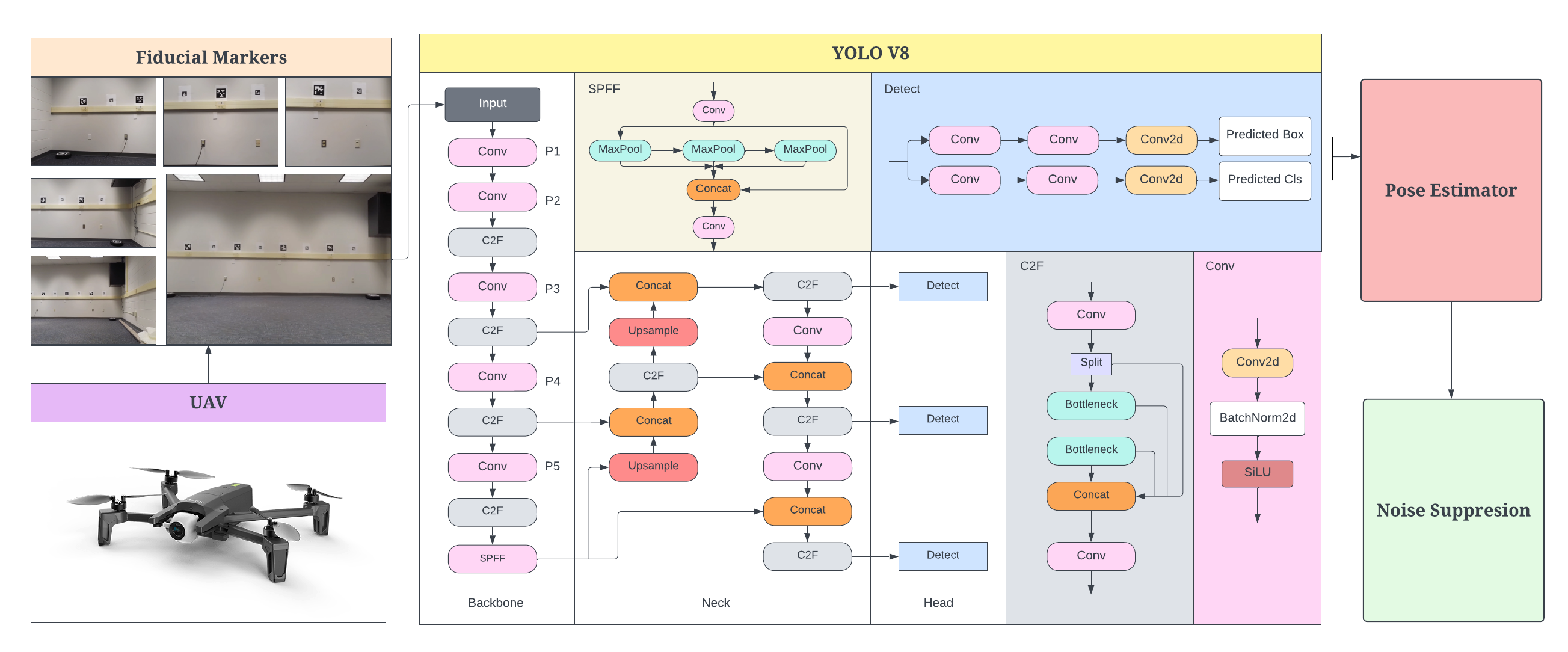}
    \caption{The UAV's onboard camera captures images of the fiducial markers. YOLO V8 identifies these markers in the images, extracting their corner points for a pose estimator to calculate the drone's position. Subsequently, a noise suppression module refines the poses to precisely establish the UAV's state.}
    \label{fig:yolov8}
\end{figure}
\vspace{-12pt}
In this work, we aim to compute a UAV's state from monocular images captured by its on-board camera. These images contain fiducial markers.
Fig.~\ref{fig:yolov8} illustrates YOLO v8-based realtime fiducial marker detection architecture which is composed of three primary components: the Backbone layer, Neck layer, and Head layer. 

The Backbone forms the base feature extractor of the model. It utilizes a modified CSPDarkNet-53 network to improve gradient flow. This backbone performs 5 downsampling operations on the input image to create a multi-scale feature pyramid with levels P1 to P5. Smaller scales capture finer details while larger scales provide semantic and contextual information. To enhance representations, the backbone integrates lightweight C2f modules that combine low-level and high-level features from different layers. It also uses standard Conv, BN, and SiLU layers to extract spatial patterns, normalize activations, and introduce non-linearities respectively. Finally, a spatial pyramid pooling fusion module generates fixed size outputs from various input sizes.

The Neck module is focused on a key challenge of detecting small, distant objects in the context of vision-based localization. To achieve this, it adopts a dual-stream feature pyramid network combining Feature Pyramid Networks (FPN) and Path Aggregation Network(PAN) structures. The FPN takes a top-down approach, upsampling and merging deep semantically rich features with lateral outputs from the backbone at each scale. This retains high-level cues that aid small object detection. Meanwhile, the PAN takes a bottom-up approach, downsampling low-level features like colors, edges, and contours and merging them with FPN outputs at each scale. This retains fine-grained spatial details lost during backbone deepening. The complementary fusion of semantic and spatial streams provides comprehensive multi-scale feature representations.

Finally, the Head contains two decoupled branches for classification and bounding box regression. 

YOLO V8 is trained on labeled data by minimizing two different losses, i.e., classification loss and regression loss as:
\begin{equation}
    \mathcal{L}_{\text{total}} = \frac{1}{N} \left( \mathcal{L}_{\text{conf}}(p, y) + \alpha \mathcal{L}_{\text{box}}(p, \hat{p}, b, \hat{b}) \right),
\end{equation}
where $N$ is the number of markers in the image, $\alpha$ is a hyperparameter that balances the importance of classification and regression, $y$ is the ground truth label of a marker, $b$ is the ground truth bounding box,
$\hat{b}$ is the predicted bounding box, $p$ is the ground truth probability distribution over classes and $\hat{p}$ is the predicted probability distribution over classes.

The classification branch categorizes objects using binary cross-entropy loss as:
\vspace{-10pt}
\begin{equation}
    \mathcal{L}_{\text{conf}}(p, y) = -\frac{1}{N} \sum_{i=1}^{N} \left[ y_i \log(p_i) + (1 - y_i) \log(1 - p_i) \right]
\end{equation}
The regression branch predicts precise box coordinates by optimizing both distribution focal loss and CIoU loss as:
\begin{equation}\label{eqn:loss}
\mathcal{L}_{\text{box}} = \lambda_{\text{dfl}} \cdot \text{DFL}(p_i, \hat{p}_i) + \lambda_{\text{ciou}} \cdot \text{CIoU}(b_i, \hat{b}_i),
\end{equation}
where $\mathcal{L}_{\text{box}}$ is the total loss for bounding box prediction, $\hat{p}_i$ is the predicted probability for object $i$, $p_i$ is the ground truth probability for object $i$, $y_i$ is the ground truth label for the object $i$. $\text{DFL}$ is the Distribution Focal Loss function,  $\lambda_{\text{dfl}}$ is the weight factor for DFL loss, $\hat{b}_i$ is the predicted bounding box coordinates for object $i$, $b_i$ is the ground truth bounding box coordinates for object $i$, $\text{CIoU}$ is the complete IoU loss function, and $\lambda_{\text{ciou}}$ is the weight factor for CIoU loss.
The Eqn.~\eqref{eqn:loss} optimizes bounding box regression using a weighted combination of distribution focal loss for classification confidence and complete IoU loss for box coordinate error. The loss balances both classification and localization performance.

\section{4D Pose Estimation for quadrotor UAVs}

Our indoor localization system for quadrotor UAVs performs marker-based pose estimation by visually detecting unique fiducial markers distributed in the environment. This is analogous to landmark-based UAV localization where the UAV detects and recognizes visual markers in place of more generic landmarks. Particularly, we place $n$  markers around the environment in such a way that could be detected when the UAV flight path brings it into close proximity. 

Let $L$ be a set of landmarks with $n$ known markers such that $L=\{l_1, l_2, \ldots, l_n\}$, $\mathbf{z}=\{\mathbf{z}_i^j\}_{i=1}^m$ be the $m$ detected marker with four corner points of $\hat{b}_i$  such that $j=1, \ldots, 4$, and $\mathbf{x}$ be the 4D pose of a UAV such that $\mathbf{x} = \{x, y, z, \theta\}$.
The goal is to estimate $\mathbf{x}$ given $L$ and $\mathbf{z}$. This can be formulated as a maximum likelihood estimation problem as:

\begin{equation}
    \mathbf{x}^* = \argmax_\mathbf{x} p(\mathbf{x} | L, \mathbf{z}) = \argmax_\mathbf{x} p(\mathbf{z} | \mathbf{x}, L) p(\mathbf{x}),
\end{equation}
where $p(\mathbf{z} | \mathbf{x}, L)$ is the landmark measurement model, and $p(\mathbf{x})$ encodes any prior knowledge about the pose. This estimates the most likely pose given the detections and known landmarks.

The perspective projection function $\pi$ transforms 3D world coordinates into 2D image coordinates such that $\pi: \mathbb{R}^3 \to \mathbb{R}^{2}$. In our setup, $\pi$ maps a landmark  $l_i^j = \{x_i^j, y_i^j, z_i^j\}$ to the corner points of bounding boxes $\mathbf{z}_i^j$. In the pinhole camera model, this transformation can be mathematically expressed as:
\begin{equation}
    \pi(l_i^j) = K \times \left[ \frac{x_i^j}{z_i^j}, \frac{y_i^j}{z_i^j},  1 \right]^{\intercal},
\end{equation}
where the matrix $K = \begin{bmatrix} f_x & 0 & c_x \\ 0 & f_y & c_y \end{bmatrix}$ encapsulates the intrinsic parameters of the camera.  $(f_x, f_y)$ and $(c_x, c_y)$ are the focal lengths and the principal points in the $x$-axis and $y$-axis directions, respectively.  

The vector $\left[\frac{x_i^j}{z_i^j}, \frac{y_i^j}{z_i^j}, 1 \right]^{\intercal}$ represents the homogeneous coordinates of the 3D point $l_i^j$. The division by $z_i^j$ ensures that the resulting 2D coordinates are normalized by the depth of the point in the camera coordinate system.
The combined effect of the intrinsic parameters and the homogeneous coordinates transformation yields the final 2D image coordinates $\pi(l_i^j)$.
The objective of Efficient Perspective-n-Point(EPnP) algorithm is to compute $\mathbf{x}^*$ by minimizing the reprojection error as: 
\begin{equation}
   \mathbf{x}^* = \min_{R, T} \sum_{i=1}^{m} \sum_{j=1}^{4} w_{ij} \cdot \lVert \mathbf{z}_i^j- \pi(R l_i^j + T) \rVert^2, 
\end{equation}
where $R$ is a rotation matrix that computes $\theta$, $T$ is a translation vector such that $T = \left[x, y, z \right]^{\intercal}$, and $w_{ij} \in \mathbb{R}^{+}$ are optional weights for $i^{th}$ marker $j^{th}$ corner point.
EPnP refines its detection utilizing redundant and geometrically diverse markers to make localization overdetermined, reduce noise/errors, and provide greater environment coverage for reducing positional ambiguity. 
Formally, each additional independent landmark detection provides more information in the maximum likelihood estimation as:
\begin{equation}
p(\mathbf{x} | L, \mathbf{z}) \propto p(\mathbf{z}_1^j | \mathbf{x}) p(\mathbf{z}_2^j | \mathbf{x}),\ldots, p(\mathbf{z}_m^j | \mathbf{x}).
\end{equation}
With more terms, small errors in individual measurements get averaged out in the combined likelihood used for localization. This leads to better accuracy and precision with more landmarks.
\section{Noise Suppression}
The combination of YOLO V8 and the EPnP algorithm yields noisy state estimates, which lead to substantial trajectory tracking errors over time. To obtain smooth state estimates, this noise must be filtered out.
To reduce the impact of noise on the state estimate, we implemented a $n_s$-order Butterworth lowpass filter which provides a maximally flat magnitude response in the passband, enabling smooth filtering around a cutoff frequency $\omega_c$. Maximally flat magnitude response allows a smooth transition between passing and attenuating frequencies around the cutoff. This prevents ringing artifacts.
Gradual roll-off after $\omega_c$ that steadily attenuates higher frequencies where sensor noise resides.

To design a Butterworth filter, first we convert our time domain signals to frequency domain using Fast Fourier Transformation. This transformation facilitates a clearer understanding of signal characteristics and aids in subsequent filter design. The transfer function $H(s)$ in frequency domain is as follows:
\begin{equation}
   H(s) = \frac{\omega_c}{\sum_{k=0}^{n_s} a_k s^k} \label{eqn:tf}
\end{equation}
where $n_s$ represents the order of the filter. With a known coefficient $a_k$, our goal is to find a new cutoff frequency $\omega_c$ in the Butterworth polynomial $B_n(s)$ based on signal characteristics to retain true localization content while suppressing noise as:
\begin{equation}
    B_n(s) = \sum_{k=0}^n a_k \left(\frac{s}{\omega_c}\right)^k = \sum_{k=0}^n \frac{a_k}{{\omega_c}^k} s^k. \label{eqn:butter_worth}
\end{equation}

\begin{figure}[htb!]
\centering
    \begin{subfigure}{0.235\textwidth}
        \centering
        \includegraphics[ width=\linewidth]{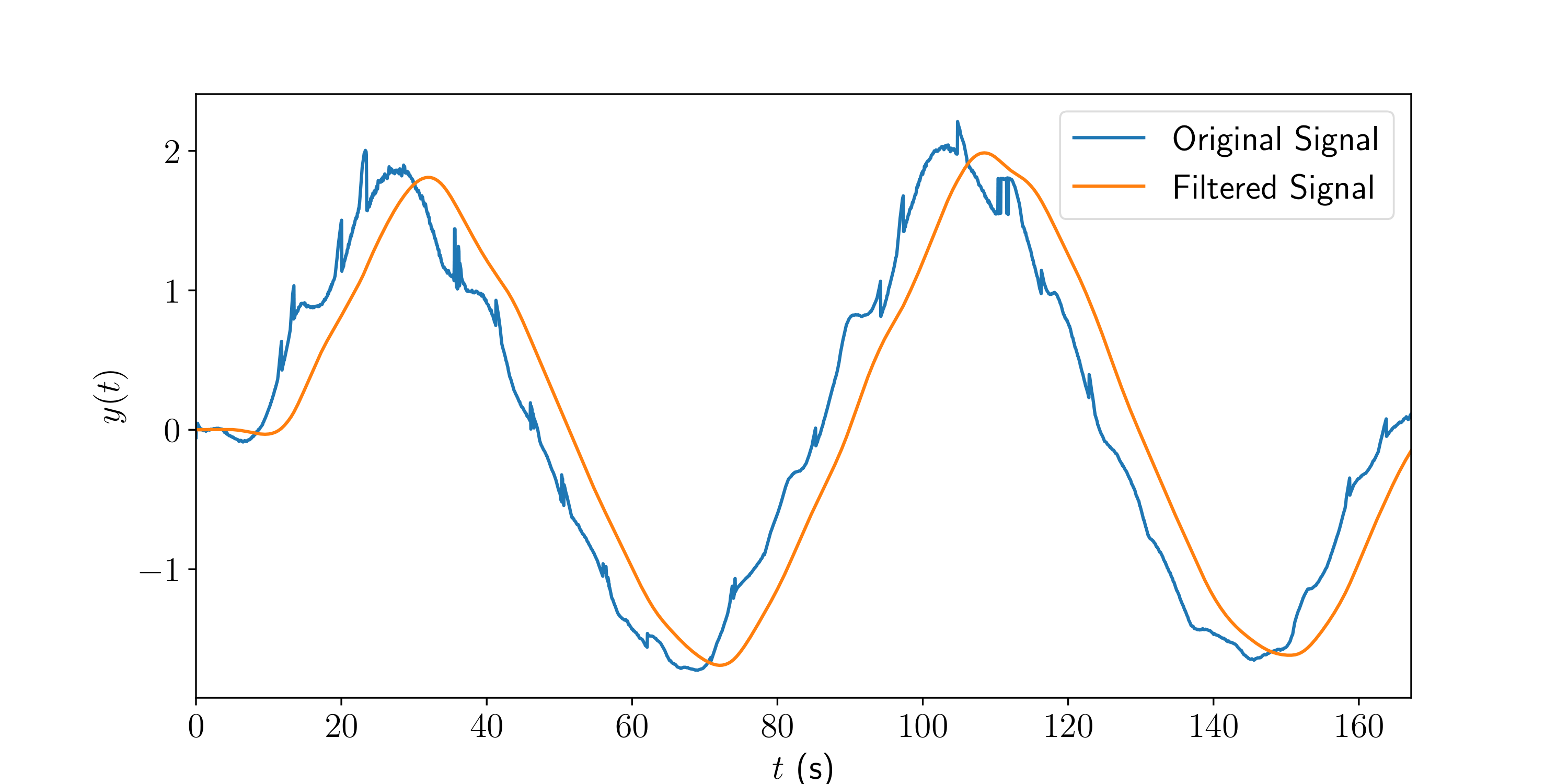}
        \caption{}
        
        \label{fig:subfig_1}
    \end{subfigure}
    ~\hfill
    \begin{subfigure}{0.235\textwidth}
        \centering
        \includegraphics[width=\linewidth]{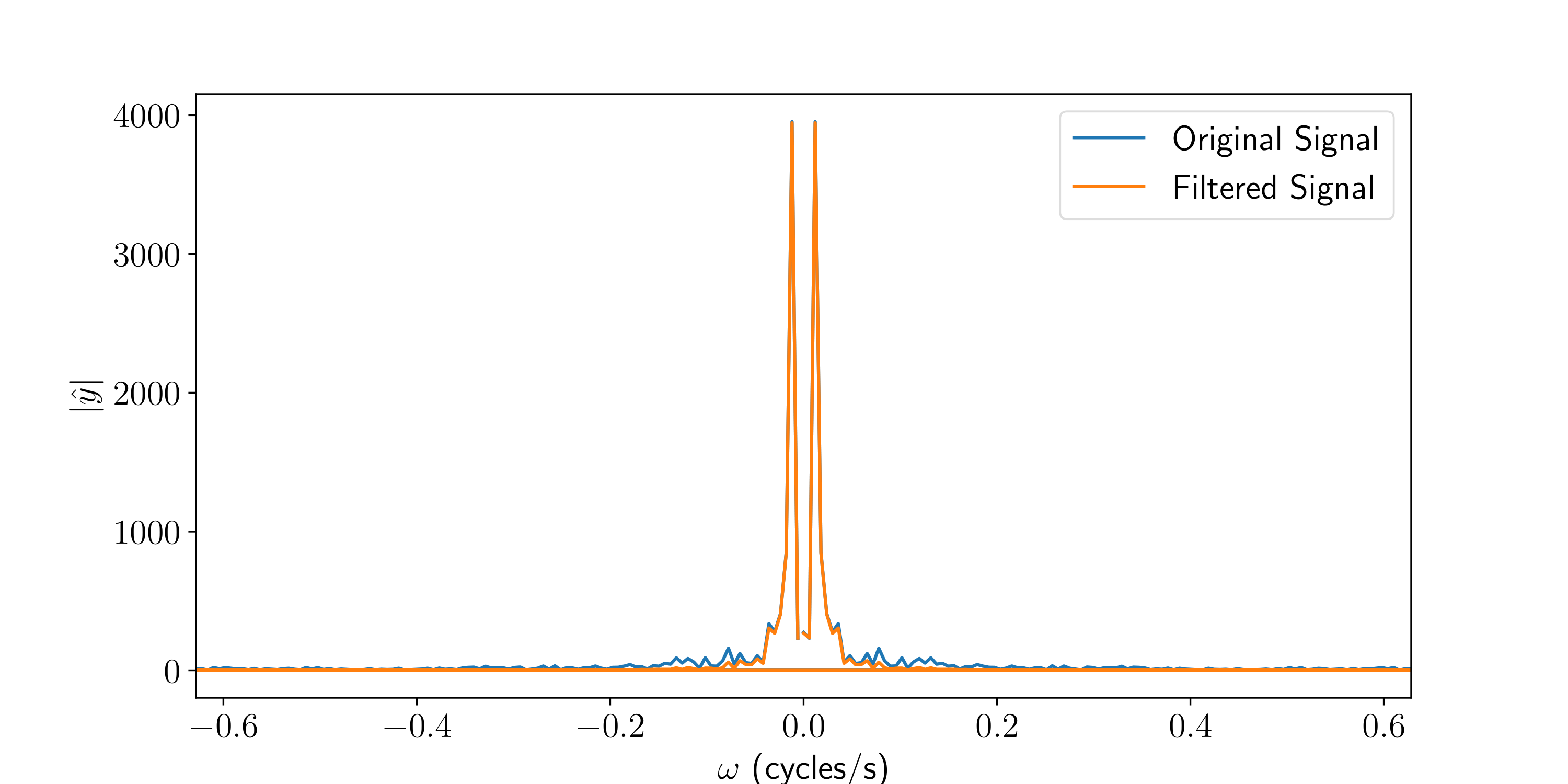}
        \caption{}
        \label{fig:subfig_2}
    \end{subfigure}

    \medskip

    \begin{subfigure}{0.235\textwidth}
        \centering
        \includegraphics[ width=\linewidth]{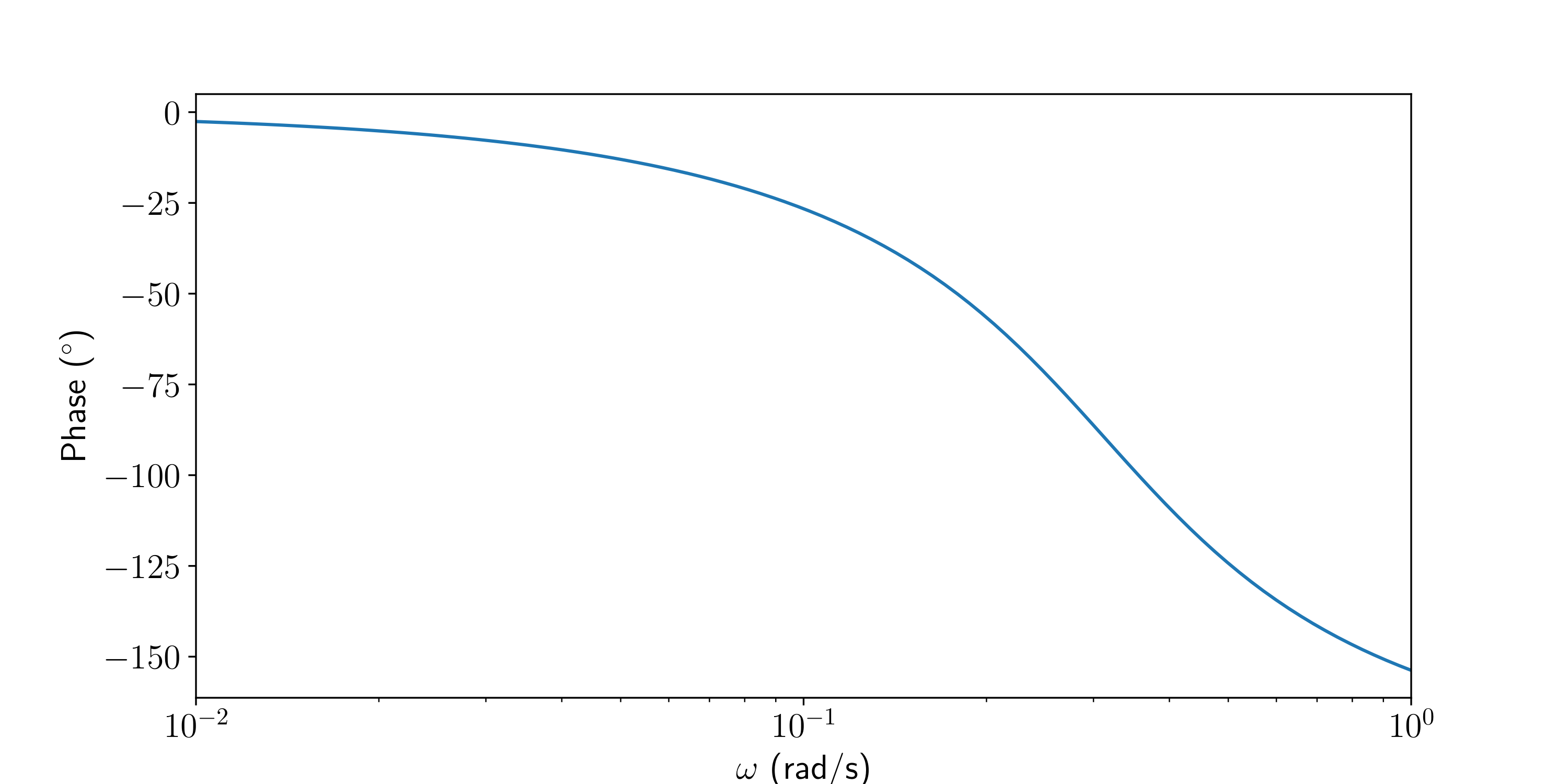}
        \caption{}
        \label{fig:subfig_3}
    \end{subfigure}
    ~\hfill
    \begin{subfigure}{0.235\textwidth}
        \centering
        \includegraphics[ width=\linewidth]{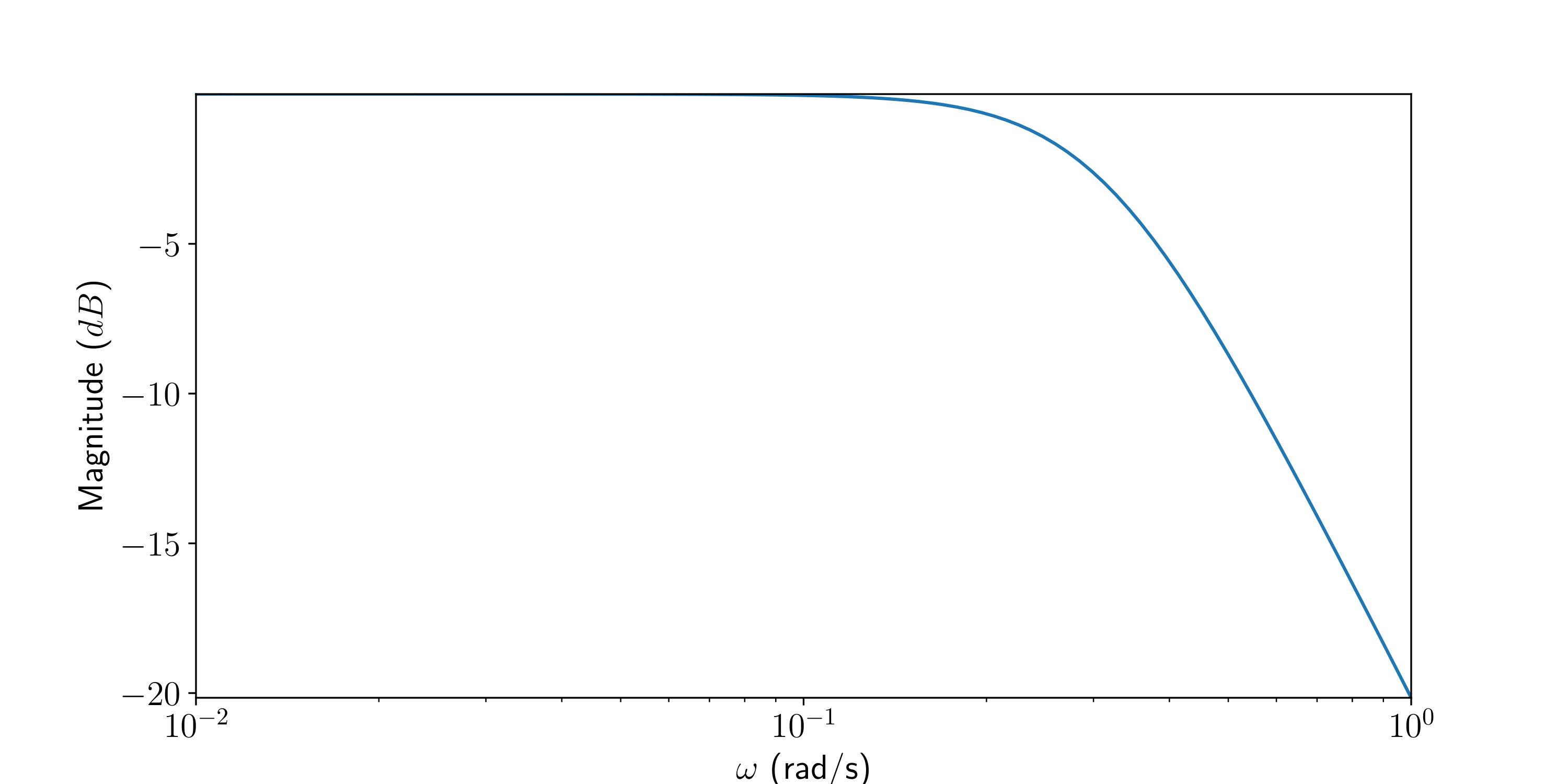}
        \caption{}
        \label{fig:subfig_4}
    \end{subfigure}

\caption{
The Blue line depicts a raw trajectory from the EPnP algorithm while the orange line depict the corresponding filtered trajectory from the Butterworth filter in Fig.~\ref{fig:subfig_1}. 
Fig.\ref{fig:subfig_2} demonstrates how a Fast Fourier Transform (FFT) is utilized to determine an appropriate cutoff frequency for the Butterworth filter. The phase response of the filter is characterized in the phase plot (Fig.~\ref{fig:subfig_3}), showing the phase shift introduced across frequencies  \textemdash specifically, a $-150^\circ$  phase shift at $1$ rad/s. Similarly, the magnitude response is shown in Fig.~\ref{fig:subfig_4}, with the filter inducing a $-20$ dB attenuation at $1$ rad/s. Analyzing these frequency responses aids in understanding the behavior of the filter and how it impacts the trajectory data.
}  
\label{fig:signal_plots}
\end{figure}

Fig.~\ref{fig:subfig_1} displays the original and filtered signals side-by-side, clearly exhibiting the noise reduction achieved, along with the resulting phase shift and decreased amplitude from the filtering process. Fig.~\ref{fig:subfig_2} reveals how the frequency components change post-filtering. Analyzing these Fourier transforms greatly assisted in selecting appropriate cutoff frequencies for the filter design. Additionally, the Bode plots in Figs \ref{fig:subfig_3} and \ref{fig:subfig_4} showcase how the filter passes certain frequencies unchanged while progressively attenuating others, providing visualization of key attributes like passband, roll-off, overall frequency response, and phase changes across frequencies. It is noteworthy that \ref{fig:subfig_3} represents the Bode phase plot and \ref{fig:subfig_4} represents the Bode magnitude plot. We obtain the phase shift from the phase plot and the amplitude from the magnitude plot in decibels. These analyses enable a comprehensive understanding of how filtering transforms signal properties through detailed time and frequency domain comparisons, coupled with characterization of the filter's impact on the signal. Insights into phase shift, amplitude reduction, cutoff frequencies, and frequency response are crucial for designing and implementing an optimal filter. The main drawback observed with the Butterworth filter is the time delay in the output signal. However, this delay is mitigated by using a 2nd-order Butterworth filter, particularly considering the lower delay inherent to this order. Furthermore, by leveraging a 30 frames per second (FPS) input frame rate and a powerful computational unit, we can effectively overcome this delay. In conclusion, this thorough analysis highlights the effectiveness of the 2nd-order Butterworth filter in smoothing trajectories and reducing noise in tag detections, underscoring its utility in enhancing data quality for trajectory analysis and related applications.
\section{Experiments}
The experiments were conducted on a desktop computer equipped with an Intel(R) Core(TM) i7-10700 CPU @ 2.90GHz and an NVIDIA Quadro P2200 GPU, boasting a GPU memory of 5 GB, along with 32 GB of RAM. The system ran on Ubuntu 20.04 LTS integrated with ROS Noetic.
For capturing ground truth trajectories, a VICON Motion Capture System was employed.
Eight Apriltag markers from the 36h11 family were positioned on the wall to serve as landmarks within our target environment. The markers ranged sequentially from 1 to 8, with each marker measuring 0.2 meters in size.
For all experiments evaluating the performance of YoloTag, a Parrot Bebop2 quadrotor-UAV was employed. The Bebop2 established a wireless connection to a desktop computer via the bebop2 autonomy ROS wrapper. Images captured by the onboard camera underwent processing to estimate the Bebop2's state. Subsequently, a PID controller facilitated real-time control of the Bebop2's motion through commands issued from the desktop computer. 
YoloTag detector is implemented using the Robot Operating System (ROS) platform. The YoloTag detector code is available at \url{https://github.com/RedwanNewaz/YoloTag}.




\subsection{Dataset Generation}
\begin{figure}[htb]
    \centering
    \begin{subfigure}{0.235\textwidth}
        \centering
        \includegraphics[width=\linewidth, height=0.8\linewidth]{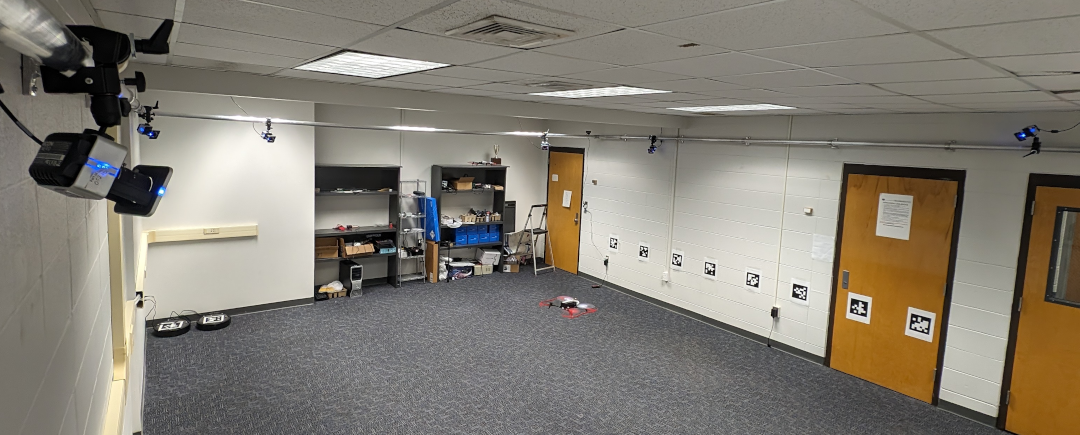}
        \caption{}
        \label{fig:vicon_setup}
    \end{subfigure}
    \begin{subfigure}{0.235\textwidth}
        \centering
        \includegraphics[width=\linewidth,height=0.8\linewidth]{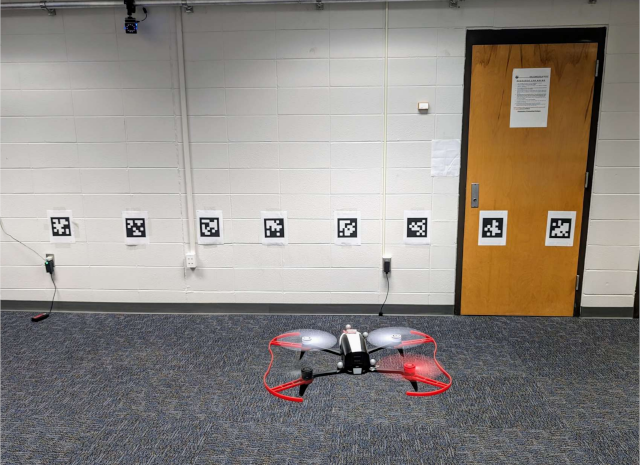}
        \caption{}
        \label{fig:bebop2_with_marker}
    \end{subfigure}
    \caption{The dataset was generated by flying a Bebop2 within a closed environment measuring $6.10$ m × $5.85$ m × $2.44$ m. A Vicon motion capture system (Fig.~\ref{fig:vicon_setup})  was employed to capture the UAV's motion, while the onboard camera was used to capture fiducial markers (Fig.~\ref{fig:bebop2_with_marker}), enabling the generation of ground truth data.}
    \vspace{-15pt}

\end{figure}

To obtain training, validation, and testing images, the Bebop2 was flown within a closed indoor environment measuring $6.10$ m × $5.85$ m × $2.44$ m, as shown in Figs.~\ref{fig:vicon_setup} and ~\ref{fig:bebop2_with_marker}.  Eight Vicon cameras were mounted on a speed rail to capture the full motion of the UAV, as depicted in Figs.~\ref{fig:vicon_setup}. Eight fiducial markers, each bearing a unique id, were affixed in static positions on the wall. The center of each marker was positioned precisely 0.724 meters from the ground. Furthermore, the distance separating any two neighboring markers measured 0.58 meters. This allowed the UAV to differentiate between the markers when detected in the camera's field of view, enabling the estimation of its state from the markers, as illustrated in Fig.~\ref{fig:bebop2_with_marker}.
Multiple experiments were conducted to generate the dataset, capturing onboard camera images at $30$ frames per second (FPS) with a resolution of $856 \times 480$ pixels. Vicon markers were used to capture the ground truth trajectories for all experiments. The training image sequences and ground truth trajectories were recorded in rosbag format.

\subsection{Training YOLO v8 with Fiducial Markers}

For training the YOLO v8 model, we utilized $9,128$ images as the training set, $5,016$ images for the validation set, and $3,148$ images for the test set.
Initial automated annotation of the fiducial markers was performed using the Apriltag detector. However, as Apriltag detector could not accurately annotate all markers in every image, manual annotation was carried out utilizing an online tool to draw bounding boxes around the missing requisite visual markers.
To achieve real-time performance, we trained a lightweight YOLO V8 model that had been pre-trained on the MS-COCO dataset. The YOLO V8 model underwent training for $10$ epochs, enabling it to classify and localize fiducial markers within image frames by generating bounding box predictions.

\subsection{Trajectory Tracking Performance}

YoloTag was designed for vision-based autonomous navigation tasks in real-world scenarios. To demonstrate its capabilities, trajectory tracking performance was compared among multiple fiducial marker detection methods. 
Two diverse trajectories were evaluated: a \textit{spiral eight} path and a \textit{rectangular eight} path.
For a spiral trajectory, where changes in direction occur at every point, prioritizing smoothness and adaptability is essential. Hence, minimum snap trajectories are preferable due to their ability to minimize jerk and provide smooth, continuous motion. In contrast, for a rectangular trajectory, where turns are sharp and changes in direction are less frequent, simplicity and predictability may be prioritized. Therefore, constant velocity trajectories are often chosen as they offer straightforward control and consistent motion along straight segments.

\begin{figure}[]
    \centering
      \begin{subfigure}[b]{0.45\linewidth}
        \includegraphics[width=1.19\linewidth, height=6cm]{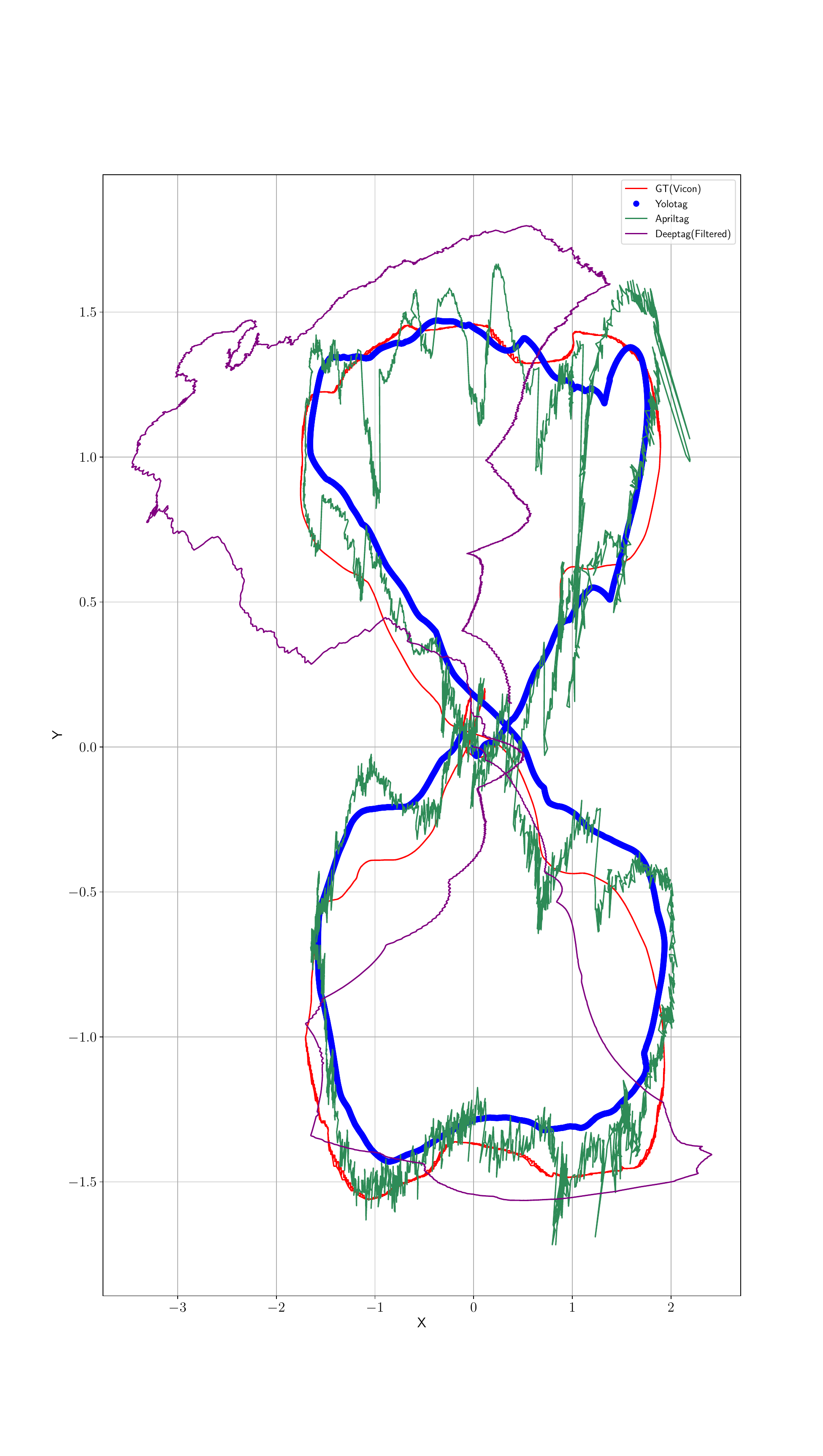}
        \vspace{-20pt}
        \caption{}
        \label{fig:Yolotag_april_spiral_comparison}
      \end{subfigure}
      \hspace{1pt}
    \centering
      \begin{subfigure}[b]{0.45\linewidth}
        \includegraphics[width=1.1\linewidth, height=6cm]{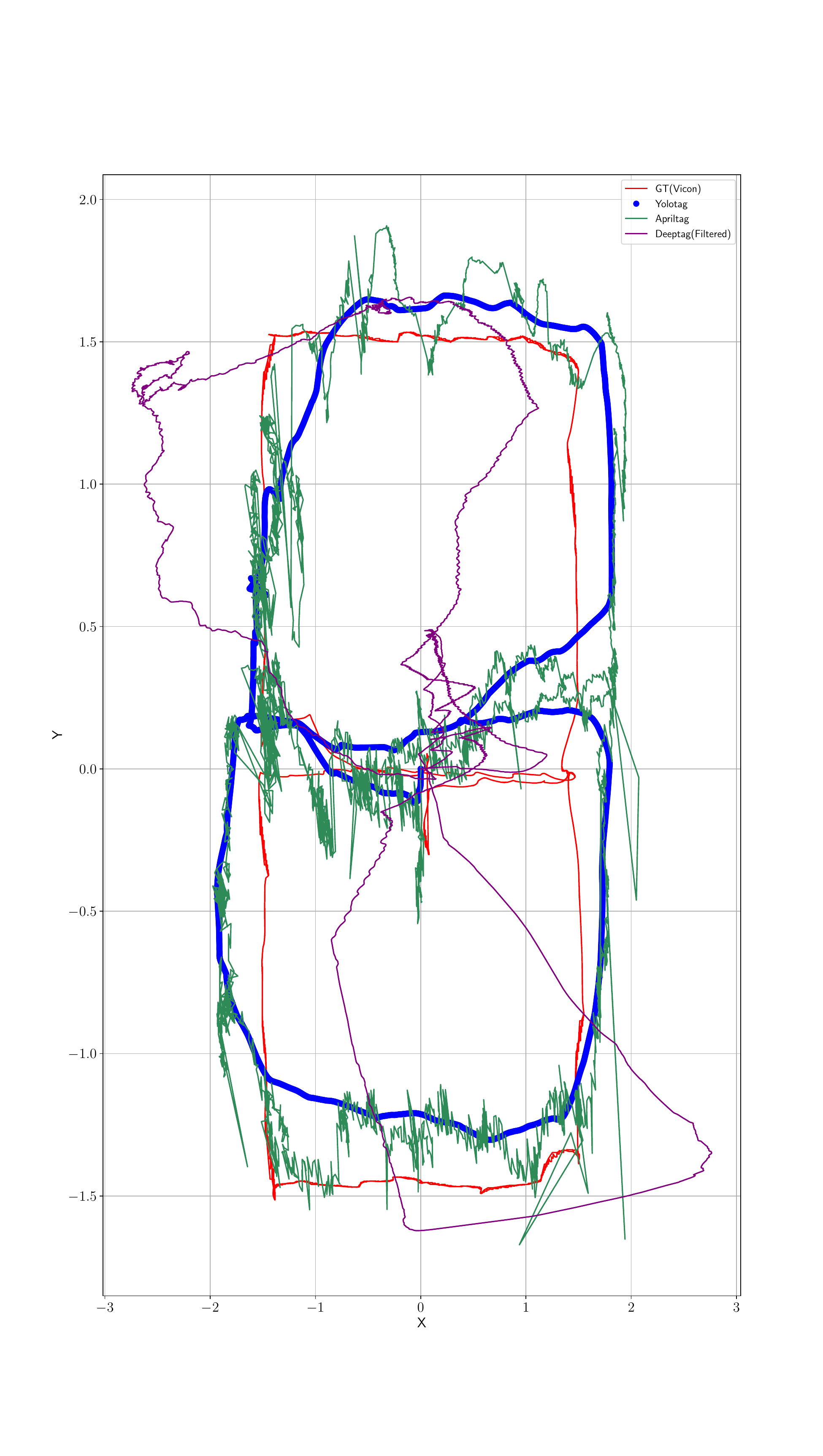}
        \vspace{-20pt}
        \caption{}
        \label{fig:Yolotag_april_rect_comparison}
      \end{subfigure}
  \caption{
  The figures show the ground truth trajectory from the Vicon system (red) compared with the raw trajectory from the AprilTag detector (green) and the filtered trajectories from the DeepTag (purple) and YoloTag (blue) detectors. Figures \ref{fig:Yolotag_april_spiral_comparison} and \ref{fig:Yolotag_april_rect_comparison} illustrate the performance of AprilTag, DeepTag, and YoloTag detectors, alongside the Vicon system, in tracking spiral and rectangular eight-shaped trajectories. The YoloTag detector outperformed AprilTag and DeepTag detectors in accurately estimating the UAV's state, demonstrating superior tracking capabilities for both trajectory profiles.
  \vspace{-22pt}
  }
  \label{fig:comparisons}
  
\end{figure}

Fig.~\ref{fig:comparisons} presents a comparison of three different trajectories generated by Apriltag, DeepTag, and YoloTag, alongside the ground truth trajectory depicted by the red line. The green line represents the trajectory obtained using Apriltag detector, while the purple line corresponds to the trajectory generated by DeepTag detector. The blue line illustrates the trajectory resulting from the proposed YoloTag detector.
As evident from Figure ~\ref{fig:comparisons}, the trajectories from other methods are noisy and deviate significantly from the ground truth trajectories. In contrast, YoloTag incorporates a second-order Butterworth filter to reduce noise in the estimated trajectory, offering advantages over DeepTag's region of interest (ROI) refinement approach. Butterworth filters excel in noise reduction, smoothing abrupt changes, and adapting to varying conditions. 
Additionally, they integrate temporal information for sequential detection and tracking of prior states, as illustrated in Fig.~\ref{fig:subfig_1}, thereby improving the accuracy and stability of the estimated trajectory.


\subsection{Benchmark}

\begin{table}[]
\vspace{12pt}
\begin{tabular}{|l|cc|cc|c|}
\hline
\textbf{Detector} & \multicolumn{2}{c|}{Hausdorff Distance} & \multicolumn{2}{c|}{Fréchet Distance} & \begin{tabular}[c]{@{}c@{}}Runtime\\ (FPS)\end{tabular} \\ \hline
 & \multicolumn{1}{c|}{\textbf{Spiral}} & \textbf{Rect.} & \multicolumn{1}{c|}{\textbf{Spiral}} & \textbf{Rect.} &  \\ \hline
YoloTag  & \multicolumn{1}{c|}{0.2464} & 0.4404 & \multicolumn{1}{c|}{2.16} & 2.064 & 55 \\ \hline
Apriltag \cite{Olson2011AprilTagAR} & \multicolumn{1}{c|}{0.386} & 0.384 & \multicolumn{1}{c|}{2.18} & 2.10 & 24 \\ \hline
DeepTag   \cite{9773975} & \multicolumn{1}{c|}{0.968} & 0.9177 & \multicolumn{1}{c|}{2.3832} & 2.276 & 03 \\ \hline
\end{tabular}
\caption{Performance Benchmark on trajectory tracking}
\label{tab:localization_benchmark}
\vspace{-20pt}
\end{table}

Table~\ref{tab:localization_benchmark} summarizes the results across three tag detection systems  \textemdash YoloTag, Apriltag, and DeepTag, evaluated on the metrics of Hausdorff Distance, Discrete Fréchet Distance, and Runtime. 
YoloTag demonstrated superior performance compared to other methods like DeepTag and AprilTag detectors in most cases, achieving superior results across all markers on both the spiral eight and rectangular eight trajectories for all three metrics.

The Hausdorff Distance quantifies the maximum deviation between two trajectories by considering the nearest point in one trajectory to a given point in the other. The Discrete Fréchet Distance quantifies the minimum length of a cord required to join the closest points along the compared trajectories. YoloTag attained the lowest Hausdorff Distance of 0.2464 for the spiral eight pattern and 0.4404 for the rectangular eight, although Apriltag detector had a marginally lower 0.384 on the rectangular eight. However, Apriltag detector's trajectory was significantly noisier than YoloTag's despite the slightly lower distance.

For Discrete Fréchet Distance, YoloTag detector again achieved the lowest values of 2.16 for spiral eight and 2.064 for rectangular eight. DeepTag detector, using Single Shot Multibox Detector (SSD) with MobileNet, exhibited subpar trajectory tracking performance with the highest Hausdorff and Discrete Fréchet Distances on both trajectories. For the Spiral Eight trajectory, DeepTag detector achieved a Hausdorff Distance of $0.968$ and a Discrete Fréchet Distance of $2.3832$. For the Rectangular trajectory, it achieved a Hausdorff Distance of $0.9177$ and a Discrete Fréchet Distance of $2.276$.

DeepTag detector's inferior performance compared to YoloTag detector is due to its reliance on the less efficient SSD with MobileNet backbone and its focus on computationally expensive per-frame tag detection without utilizing sequential image information. 
Although there is a GPU implementation of AprilTag detector$^{\footnotemark}$ 
 \footnotetext{\url{https://github.com/NVIDIA-ISAAC-ROS/isaac_ros_apriltag}} that can run faster, its performance is worse than the CPU version. 
AprilTag detector CPU version performed better than DeepTag detector but suffered from detection robustness issues. Tracking by individually estimating each tag's pose degraded performance, although combining AprilTag detection algorithm with the EPnP algorithm improved trajectory tracking. However, AprilTag detector frequently failed to detect all markers and was more susceptible to motion blur and noise than YoloTag detector. While YoloTag detector can enhance performance through training on diverse datasets, AprilTag detector is limited by its reliance on conventional image processing techniques and predefined marker designs, lacking YoloTag detector's generalizability.

In runtime benchmarking, the YoloTag detector, which is GPU-accelerated and Python-based, outperformed both the AprilTag detector, which is CPU-based and written in C++, and the DeepTag detector, which is also GPU-accelerated and Python-based. Leveraging YOLOv8's lightweight architecture, YoloTag detector achieved 55 FPS on an NVIDIA Quadro P2200 GPU, while AprilTag detector was limited to 24 FPS for real-time tracking. 
 DeepTag detector only managed 3 FPS when processing the 8 markers required for navigation, making its complex pipeline unsuitable for such tasks. This substantial YoloTag detector performance advantage over Apriltag and DeepTag detectors in localization accuracy metrics like Hausdorff and Discrete Fréchet Distances, as well as processing speed, makes YoloTag detector the most suitable choice for robust and efficient multi-marker trajectory tracking for navigation tasks.

\section{Conclusion}

This work introduces YoloTag, a novel fiducial marker detection architecture designed for vision-based UAV navigation tasks. YoloTag employs a lightweight YOLO v8 object detector for accurate marker detection and an efficient multi-marker based pose estimation algorithm to robustly compute the UAV's pose across diverse real-world conditions.
A lightweight YOLO v8 model is trained on a dataset generated from real-world experiments involving a UAV capturing onboard camera images. A mixed annotation approach, combining an Apriltag detector and manual annotation, enables an efficient training regime.
Notably, by incorporating sequential pose information through a second-order Butterworth filter, YoloTag achieves superior performance and real-time capabilities compared to existing methods, as evidenced by its performance across multiple distance metrics. The proposed architecture offers a robust, efficient, and real-time solution for marker-based UAV localization and navigation in GPS-denied environments, outperforming traditional approaches.
Future work will focus on employing YoloTag for UAV localization by detecting common objects, thereby overcoming the limitation of fiducial detection methods that are constrained to predefined markers.

\bibliographystyle{ieeetr}
\bibliography{ref}

\end{document}